\documentclass[10pt, conference, letterpaper]{IEEEtran}

\makeatletter
\def\ps@headings{%
\def\@oddhead{\mbox{}\scriptsize\rightmark \hfil \thepage}%
\def\@evenhead{\scriptsize\thepage \hfil \leftmark\mbox{}}%
\def\@oddfoot{}%
\def\@evenfoot{}}
\makeatother
\usepackage[left=0.655in, right=0.655in, top=0.75in, bottom=1.0in]{geometry}
\usepackage[utf8]{inputenc}
\usepackage{graphicx}
\usepackage{verbatim}
\usepackage{hyphenat}
\usepackage{epstopdf}
\usepackage{graphics}
\usepackage{amsmath}
\usepackage{amsfonts}
\usepackage{tabularx}
\usepackage{enumerate}
\usepackage{xcolor}
\usepackage{subcaption}
\usepackage[normalem]{ulem}

\usepackage{siunitx}
\usepackage{booktabs}
\usepackage{arydshln}   % provides \hdashline
\usepackage{xspace}
\newcommand{\GEE}{GEE\xspace}

\title{\GEE: A Gradient-based Explainable Variational Autoencoder for Network Anomaly Detection}

\newcommand{\phong}{0mm}
\newcommand{\bryan}{-25mm}
\newcommand{\karwai}{0mm}
\newcommand{\mc}{-30mm}
\newcommand{\dinil}{0mm}

\author{\IEEEauthorblockN{\hspace{-\phong}Quoc Phong Nguyen\hspace{\phong}}
\IEEEauthorblockA{\hspace{-\phong}National University of Singapore\hspace{\phong} \\
\hspace{-\phong}qphong@comp.nus.edu.sg\hspace{\phong}}
\IEEEauthorblockN{\\\hspace{-\bryan}Kian Hsiang Low\hspace{\bryan}}
\IEEEauthorblockA{\hspace{-\bryan}National University of Singapore\hspace{\bryan} \\
\hspace{-\bryan}lowkh@comp.nus.edu.sg\hspace{\bryan}}
\vspace{-7mm}
\and
\IEEEauthorblockN{\hspace{-\karwai}Kar Wai Lim\hspace{\karwai}}
\IEEEauthorblockA{\hspace{-\karwai}National University of Singapore\hspace{\karwai} \\
\hspace{-\karwai}karwai.lim@nus.edu.sg\hspace{\karwai}}
\IEEEauthorblockN{\\\hspace{-\mc}Mun Choon Chan\hspace{\mc}}
\IEEEauthorblockA{\hspace{-\mc}National University of Singapore\hspace{\mc} \\
\hspace{-\mc}chanmc@comp.nus.edu.sg\hspace{\mc}}
\vspace{-7mm}
\and
\IEEEauthorblockN{\hspace{-\dinil}Dinil Mon Divakaran\hspace{\dinil}}
\IEEEauthorblockA{\hspace{-\dinil}Trustwave\hspace{\dinil} \\
\hspace{-\dinil}dinil.divakaran@trustwave.com\hspace{\dinil}}
}

\begin{document}

\maketitle

\begin{abstract}
This paper looks into the problem of detecting network anomalies by analyzing NetFlow records. % Evolving network characteristics, lossy information in NetFlow records, lack of clean labeled data, etc. make this problem challenging. 
While many previous works have used statistical models and machine learning techniques in a supervised way, such solutions have the limitations that they require large amount of labeled data for training and are unlikely to detect zero-day attacks. Existing anomaly detection solutions also do not provide an easy way to explain or identify attacks in the anomalous traffic. To address these limitations, we develop and present \GEE, a framework for detecting and explaining anomalies in network traffic. \GEE comprises of two components: (i)~Variational Autoencoder (VAE) --- an unsupervised deep-learning technique for detecting anomalies, and (ii)~a gradient-based fingerprinting technique for explaining anomalies. Evaluation of \GEE on the recent UGR dataset demonstrates that our approach is effective in detecting different anomalies as well as identifying fingerprints that are good representations of these various attacks. 
\end{abstract}

\begin{IEEEkeywords}
Anomaly Detection, NetFlow Records, Gradient-based Fingerprinting
\end{IEEEkeywords}

\section{Introduction}
\label{sec:intro}

Anomalies in network can arise due to attacks and threats in the cyberspace, such as different kinds of DDoS attacks (e.g., TCP SYN flooding, DNS amplification attacks, etc.), brute force attempts, botnet communications, spam campaign, network/port scans, etc.  Network anomalies may also manifest due to non-malicious causes, such as faults in network, misconfigurations, BGP policy updates, changes in user behaviors, etc. Detection of anomalies, possibly in real time, is of utmost importance in the context of network security, in particular as we are currently witnessing  a continuous change in the threat landscape, leading to increase in the intensity, sophistication and types of attacks~\cite{symantec-report-2018}. This trend is expected to continue as the IoT market keeps expanding. Indeed, cyber criminals now target to harness more resources by exploiting IoT devices that are, both, likely to be much more vulnerable than typical computers
%(as they are often left unpatched and unattended), 
and huge in number (e.g., Mirai attack~\cite{Understanding-Mira-2017}). 

In this paper, we look into the problem of detecting anomalies in large-scale networks, like that of an Internet Service Provider (ISP). While the problem of network anomaly detection has been of keen interest to the research community for many years now, it still remains a challenge for a number of reasons.
First, the characteristics of network data depend on a number of factors, such as end-user behavior, customer businesses (e.g., banking, retail), applications, location, time of the day, and are expected to evolve with time. Such diversity and dynamism limits the utility of rule-based detection systems. 

Next, as capturing, storing and processing raw traffic from such high capacity networks is not practical, Internet routers today have the capability to extract and export meta data such as NetFlow records~\cite{RFC7011}. With NetFlow, the amount of information captured is brought down by orders of magnitude (in comparison to raw packet capture), not only because a NetFlow record represents meta data of a set of related packets, but also because NetFlow records are often generated from sampled packets. 
Yet, NetFlow records collected from a modest edge router with 24x10Gbps links for a 5-minute interval can easily reach a few GBs. 
% (depending on NetFlow configurations and connection size distribution).} \todo{should move to somewhere else?}
% The concerns on privacy is also significantly reduced as there is no payload captured; besides, client IP addresses can be anonymized to further address the privacy concerns (e.g, using prefix-preserving anonymization~\cite{prefix-preserving-anony-ICNP-2002}). 
However, with NetFlow, useful information such as suspicious keywords in payloads, TCP state transitions, TLS connection handshake, etc.\ are lost; indeed with sampled NetFlow, even sizes of each packet, time between consecutive packets, etc.\ are unavailable. Therefore, anomaly detection solutions have to deal with lossy information. 

% Capturing, storing and processing raw traffic capture from such high capacity networks is not practical due to scalability issues. Instead, Internet routers today have the capability to extract and export meta data such as NetFlow records~\cite{RFC7011}. With NetFlow, the amount of information captured is brought down by orders of magnitude (in comparison to raw packet capture), not only because a NetFlow record represents meta data of a collection of related packets, but also because NetFlow records are often collected from sampled packets. The concerns on privacy is also significantly reduced as there is no payload captured; besides, client IP addresses can be anonymized to further address the privacy concerns (e.g, using prefix-preserving anonymization~\cite{prefix-preserving-anony-ICNP-2002}). 

Finally, the SOC (security operation center) analysts have a limited budget, within which they have to analyze the alerts raised by an anomaly detector, for purposes such as alert escalation, threat and attack mitigation, intelligence gathering, forensic analysis, etc. Therefore, anomaly detectors should go beyond merely indicating the presence of anomalies; the time of anomaly, its type, and the corresponding set of suspicious flows are in particular very useful for further analysis. In general, the more the information that can be passed (along with the alerts) to the analysts, the easier is the task of analysis and quicker the decision process. 

%Although machine learning techniques are attractive, lack of labeled data for training models poses another serious challenge.
One way to address the above challenges is to apply statistical models and machine learning algorithms. Considering anomaly detection as a binary classification problem, a supervised machine learning model can be built using normal and anomalous data, for classifying anomalies. However, existing approaches have the following limitations. First, many of them exploit only a small number of features (e.g. traffic volume, flow rates, or entropy) \cite{characterization-nw-anomalies-2004, Astute-SIGCOMM-2010, AD-time-varying-nw-TCNS-2015}. Such approaches require the users to apply domain knowledge to select the right set of features which may not always be feasible and optimal. Second, supervised approaches require large sets of data with ground truth for training. 
Note that as the network characteristics and attacks evolve, models have to be retrained and the corresponding labeled datasets have to be made available. This requires costly and laborious manual efforts, and yet, given the size of traffic flowing through backbone network, it is highly impractical to assume all data records to be correctly labeled~\cite{ML-for-NID-SP-2010}. Besides, supervised approaches are unlikely to detect unknown and zero-day attack traffic.

To address these limitations, in this paper, we develop and present \GEE, a \underline{g}radient-based \underline{e}xplainable variational auto\underline{e}ncoder, for detecting as well as explaining anomalies in large-scale networks. \GEE comprises of two important components: (i)~a variational autoencoder (VAE), an unsupervised, deep-learning technique for detecting anomalies in network traffic; and (ii)~a gradient-based fingerprinting technique for explaining threats and attacks in anomalous traffic. \GEE addresses the limitations of existing approaches in the following way. First, modeling with VAE allows the use of a large number of features from the dataset thus relieving the need to employ domain knowledge to select the ``right'' set of features. Second, \GEE works with unlabeled NetFlow traces which is the norm in ISP networks for data collection. Finally, \GEE provides explanation as to why an anomaly is detected.

% \todo{should we mention about online training, and streaming capability?}

To summarize, the contributions of this paper are as follow:
\begin{itemize}
    \item We present a VAE based framework for network anomaly detection that is scalable in, both, the size of data and the feature dimension. 
    %\todo{large number of data points with high dimensional features? -> scalable in both the number of dimensions and the size of the data}
    
    \item We develop a gradient-based  framework to explain the detected anomalies, and identify the main features that cause the anomalies. This is of great impact in practice as deep  learning techniques are notorious for  non-interpretability. 
    \item Our framework \GEE makes it possible to identify network attacks using \emph{gradient-based fingerprints} generated by the VAE algorithm. To the best of our knowledge, this is the first attempt to explain network anomalies using gradients generated from a VAE model.
\end{itemize}
Evaluation using the UGR dataset~\cite{UGR16} shows that \GEE is effective in identifying the network attacks present, such as Spam, Botnet, (low-rate) DoS and port scans. Using \GEE, we can identify the features that define these attacks and the fingerprints derived (when ground truth is available) can be utilized to further improve the detection accuracy.    
    
%This paper starts with the related work in Section~\ref{sec:related}. Next, 
We provide an overview of the deep learning model in Sections~\ref{sec:model}; and in Section~\ref{sec:method}, we present our anomaly detection framework \GEE. Finally, we evaluate \GEE in Section~\ref{sec:eval}.

\section{Related Work}
\label{sec:related}

% \todo{MC - the related work is too long, I am shortening it ... }

Detection of network anomalies (such as threats, intrusions, attacks, etc.) is a widely and deeply studied research problem. Past works have developed solutions based on varying approches, for example, rule-based systems~\cite{rule-based-INFOCOM-2009}, information theoretic techniques~\cite{max-entropy-est-AD-IMC-2005, entropy-based-anomaly-detection-IMC-2008}, signal analysis~\cite{wavelet-analysis-IMW-2002}, statistical models and hypothesis testing~\cite{alpha-stable-model-TDSC-2011,NADA-TON-2018}, as well as data mining and machine learning algorithms~\cite{FIM-AD-2012, Disclosure-ACSAC-2012}. As computational resources becoming cheaper, there has been an increasing interest in applying machine learning techniques for detecting anomalies in network. In the following, we present an overview of machine learning techniques applied on NetFlow data (or other forms of aggregate data usually collected from routers in ISP networks), for the purpose of detecting anomalies.

%\todo{Our definition of anomalies?}
%\subsection{Machine Learning Techniques for Network Anomaly Detection}

PCA (Principal Component Analysis) has been used to separate traffic space into two different subspaces (`normal' and `anomalous') for anomaly detection. Aggregate byte counts of links in an ISP network was initially used as a feature~\cite{PCA-link-stats-AD-CCR-2004}; and in Lakhina et al.~\cite{characterization-nw-anomalies-2004}, the solution was extended to work on more granular data.
% ---byte and packet counts of flows identified using the common 5-tuple of IP source and destination address, source and destination ports and protocol. The basic idea was to project data into a small and fixed number of principal components, where the first few components correspond to normal subspace and the remaining correspond to anomalous subspace. The projection on to the anomalous subspace is analyzed (for abnormal changes) and a decision criterion applied (using a hypothesis test) to subsequently detect anomalies. 
An inherent disadvantage of PCA is that the components in the reduced subspace do not provide an interpretation of features in the original space. Besides, as later studies have shown~\cite{sensitivity-PCA-AD-SIGMETRICS-2007, PCA-probs-sols-INFOCOM-2009}, there are other challenges, an important one being that PCA-based detection solution is highly sensitive to the dimension of normal subspace and the detection threshold, both of which are not easily determined. 

Another approach to detect anomalies is to first model the normal traffic, and then use a statistical decision theoretic framework to detect deviation from normal data. For example, Simmross-Wattenberg et al.~\cite{alpha-stable-model-TDSC-2011} used $\alpha$-stable distributions to model 30-minute time bins of aggregate traffic rate, and generalized likelihood ratio test 
% (GLRT) 
for classifying whether the time windows are anomalous. Since a number of parameters have to be learned, large amounts of labeled data have to be continuously made available for the model to adapt with time. 

%Though limited to detection of botnet servers, one of the interesting works that applied supervised machine learning on NetFlow records is 

In Bilge et al.~\cite{Disclosure-ACSAC-2012}, a Random Forests classifier was trained to detect C\&C servers in a supervised way, %~\cite{RF-ML-2001} 
using features extracted from labeled NetFlow records.
%Since the goal was to detect botnet servers, the authors also developed and applied a heuristic to restrict analysis to only NetFlow records involving servers. 
To reduce false positive rate, however, the system relies on reputation systems such as malicious domain classification system, Google Safe Browsing, etc.
%to compute a reputation score. 
This affects detection accuracy since most recent botnets and evasive botnets may not have a low reputation score. 

% \todo{DD: can we say our method can do incremental learning?} 
%\subsection{Deep Learning}

More recently, deep neural network models (also generally referred to as deep learning models) have been embraced by the research community to tackle anomaly detection problem in networking setting.  
% Some existing works employ supervised deep learning models to classify for anomalies or cyberattacks, by making use of labeled information that is made available from manual tagging or simulated ground truth. 
Existing supervised deep learning approaches include work by Tang et al.~\cite{TangMhamdiMcLernonZaidiGhogho2016} that
utilizes a classical deep neural network for flow-based anomalies 
classification in a Software Defined Networking (SDN) environment, and use of the recurrent neural network (RNN) models for developing intrusion detection solution~\cite{TorresCataniaGarciaGarino2016}. %\cite{KimKimThuKim2016}. 
% and system call traces~\cite{KimYiLeePaekYoon2016}.
% The RNN models work by modelling the data as a sequence of information using the long short-term memory (LSTM) unit that is made popular from natural language processing applications. 

% Instead of feeding the data directly into a supervised deep learning model for classifying anomalies,
There are recent works that use unsupervised deep learning
models to transform the data into lower rank features
before applying supervised machine learning. 
Several prior approaches first employ autoencoders~\cite{Baldi2012} and their variants to extract the compressed latent representation as features, and subsequently use these features for anomaly detection by training standard classifiers such as 
% softmax regression~\cite{JavaidNiyazSunAlam2016} and
Random Forests~\cite{ShoneNgocPhaiShi2018}. 
% Comparing to original features, the features generated 
% through autoencoders have also been shown to improve 
% the performance of Gaussian Naive Bayes, SVM, and Xgboost for anomaly detection~\cite{YousefiAzarVaradharajanHameyTupakula2017}.
% Different to the above, Cao et al.~\cite{CaoNicolauMcDermott2016}
% use the extracted features to train a density estimation model
% which does not require labeled data, albeit the ground
% truth information is used to separate the anomalies from
% the training set.
% In addition to autoencoders, Alom and Taha~\cite{AlomTaha2017} 
% consider also the Restricted Boltzmann Machine (RBM)
% for feature extraction and dimensionality reduction
% before clustering with iterative k-means. Their results
% indicated that the reduced latent space representation 
% is useful. 
% One limitation with the above methods is that the
% deep learning models are utilized in a two-stage process
% where the labeled information or results are not used to
% fine-tune the generated features.

Anomaly detection methods that are solely based on unsupervised
deep learning models have also been experimented. These 
models do not require labeled information and instead
exploit the fact that anomalous behaviors tend to differ greatly from the standard or normal behavior of the network. 
% It is generally assumed that suspicious activities take up
% only a small portion of the data, and can be identified
% when compared against the normal behavior learned from
% the majority of the data. 
Fiore et al.~\cite{FiorePalmieriCastiglioneDeSantis2013}
made use of discriminative restricted Boltzmann machine (RBM) for anomaly detection
on network data; while 
Mirsky et al.~\cite{MirskyDoitshmanEloviciShabtai2018}
proposed an ensembles of light-weight autoencoders
for real time network intrusion detection, although their focus is on scalability of the system.
% demonstrated through Raspberry PI in an IoT network.
Further, An and Cho~\cite{AnCho2015} 
demonstrated that the
VAE performs much better than AE and PCA
on handwritten digit recognition
and network intrusion detection. 
However, the VAE model was trained using data labeled as normal, i.e., the anomalies are 
removed from training, which is difficult to do in practice.
The above~\cite{FiorePalmieriCastiglioneDeSantis2013, MirskyDoitshmanEloviciShabtai2018, AnCho2015}
are also known as semi-supervised learning.
% since the anomalous data (as indicated by the labels)
% is removed during training. 
% The reason for this is
% that an expressive model can potentially overfit the
% data and hence treat the anomalies as normal.

% \cite{TuorKaplanHutchinsonNicholsRobinson2017}
% use RNN for unsupervised anomaly detection on system logs
% to look for insider threats.

Different from the above, we develop an anomaly detector using an unsupervised deep learning technique without using labeled information. While existing works (e.g.,~\cite{AnCho2015}) stop at the task of anomaly detection, 
% and do not provide explanation as to why an anomaly is detected, 
in our work, we also provide a gradient-based technique for explaining why the anomalies are detected, together with their relevant features.
% enable the analysts to understand the anomalies better. 

\section{Unsupervised Deep Learning Models}
\label{sec:model}

% \todo{a brief intro to big data and deep learning here would be useful ... }

Given large amount of data being constantly collected by the network routers, existing anomaly detection
frameworks tend to employ simple methods such as threshold-based
approaches or simply PCA for scalability reason. 
Within the past decade, however, we see a rise of application
of deep learning models due to their ability to handle
big datasets as well as to train real-time in a streaming manner. This is while retaining their state-of-the-art
performance in various tasks like real time object 
recognition~\cite{RenHeGirshickSun2015} and 
fraud detection~\cite{RoySunMahoneyAlonziAdamsBeling2018}.

Additionally, deep learning models like the 
\emph{variational autoencoder} (VAE) are shown to be robust to 
noisy data~\cite{HsuZhangGlass2017}, and thus especially
suitable for modeling network flows which are very noisy 
in nature.
Although deep learning models are often criticized 
for their lack of interpretability, recent advances 
have brought forward better understanding of these models,
in particular, attributing the causes to the 
results~\cite{AnconaCeoliniOztireliGross2018}.

We focus on the VAE, a probabilistic 
generalization of the AE, for
anomaly detection on network data. 
Note that
the VAE has been shown to be more flexible and 
robust~\cite{AnCho2015} compared to the AE.
Further, we demonstrate how we can use gradient 
information from the VAE for interpretation purpose.
For instance, it can be used to analyze how a certain set
of features is more likely to explain a certain anomaly. 
In the following, we first describe the AE model 
since the VAE has the same deep architecture as 
the AE. 
% We then discuss their differences, in
% particular, the treatment of the latent representation
% variables.

\subsection{Autoencoder (AE)}

\begin{figure}[t]
    \begin{center}
    % \vskip 0.02in
        \includegraphics[width=0.9\columnwidth]{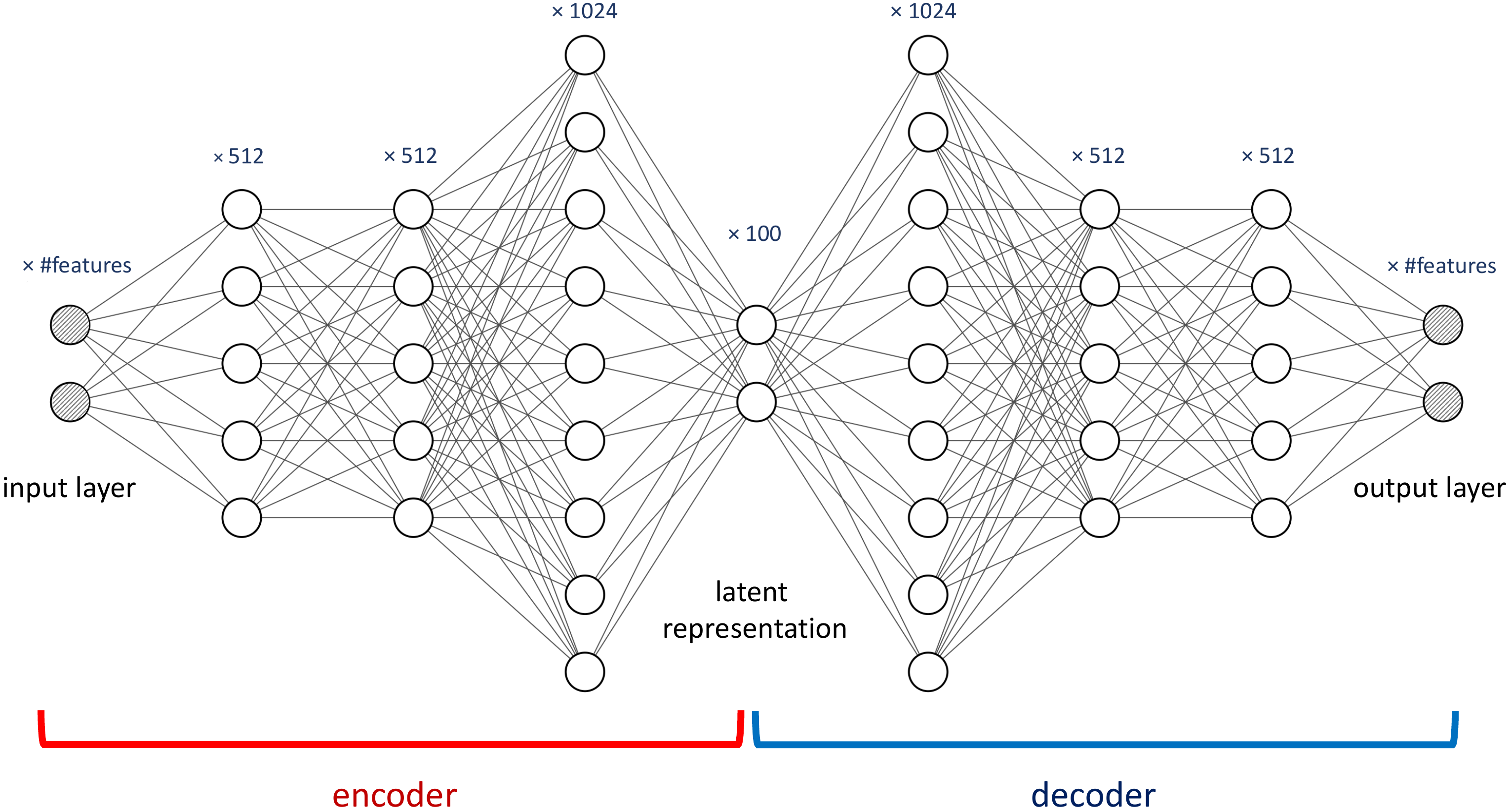}
        \caption{
            The autoencoder (AE and VAE) architecture. 
            % Separated by the latent representation layer
            % with dimensionality of 100, 
            % the region on the left side
            % corresponds to the deep encoder while the other 
            % illustrates the deep decoder. In AE, features
            % are passed from the left as inputs to be encoded into their
            % latent representation, and subsequently used to
            % generate a reconstruction of the features 
            % by the decoder as outputs. 
            % Nodes that are shaded represent the observed data
            % (used as both inputs and outputs), 
            % while the unshaded nodes are unobserved latent
            % variables corresponding to the hidden layers. 
            % The exact size or dimension of the 
            % layers is shown above the nodes.
            % Note that our variational autoencoder (VAE) model
            % employs the same architecture with the AE.
        }
        \label{fig:autoencoder}
    \end{center}
    \vskip -0.2in
\end{figure}

An AE is made of three main layers which correspond to
(i) the input layer to take in the features, 
(ii) the latent representation layer of the features,
and (iii) the output layer which is the reconstruction
of the features. 
% Note that the latent representation layer is a hidden
% layer since its values are unknown and need to be learned.
The AE consists of two parts called encoder and decoder 
respectively. The encoder maps the input into its
latent representation while the decoder attempts to 
reconstruct the features back from the latent representation.
The encoder may be deep in the sense
that information from the input is passed through several
mappings (and hidden layers) similar to the deep architecture
in a supervised deep learning model; likewise for the decoder.
% Although traditionally the dimension of the latent representation is chosen to be much smaller than the number of
% features, it is not necessarily the case in practice.
% This is known as being overcomplete when 
% the size of a hidden layer is much greater than the 
% input dimension, and it has been successfully applied 
% to structured prediction of 3D poses and denoising
% \cite{VincentLarochelleBengioManzagol2008}
% \cite{TekinKatirciogluSalzmannLepetitFua2016}.
In this paper, we set the size of the latent 
representation layer to be 100. 
% \todo{compared with 53 dimension input}. 
In addition, the encoder and 
the decoder each have three hidden layers with size 512, 512, 
and 1,024 respectively, as illustrated in Fig.~\ref{fig:autoencoder}.
In the figure, nodes that are shaded represent the observed data
(used as both inputs and outputs), 
while the unshaded nodes are unobserved latent
variables corresponding to the hidden layers. 
The exact size or dimension of the 
layers is shown above the nodes.

The links between the layers show how the values of the
\emph{next} layer can be computed. Commonly, the 
value of one hidden layer $\vec{h}_{i}$ can be computed as 
\begin{align}
    \vec{h}_{i} = g( \mathbf{W}_{i} \, \vec{h}_{i-1} + \vec{b}_{i} )
\label{eq:AE_computation}
\end{align}
where $\vec{h}_{i-1}$ is a vector of values for the previous
layer, $\mathbf{W}_{i}$ is a matrix of weights that signifies
the relationship from the previous layer, and $\vec{b}_{i}$ is 
a vector of bias terms. 
Both $\mathbf{W}_{i}$ and $\vec{b}_{i}$ are parameters to be learned 
through model training. Here, $g()$ is known as the 
\emph{activation function} that transforms the computation
in a non-linear way and allows complex relationship to be learned.
Popularly used activation functions include the \emph{sigmoid} function 
$g(x) = (1 + e^{-x})^{-1}$ and the rectified linear unit (ReLU)
$g(x) = \max(0,x)$, which will be used in this paper.
% Replicating the above computation for all nodes (from left to right as in Fig.~\ref{fig:autoencoder})
% constitutes the full process of encoding and decoding.
The learning of the parameters are generally achieved
by minimizing the reconstruction errors 
(e.g., mean square errors) via
backpropagation with random initialization, and can be
optimized with a variety of optimizers such as stochastic
gradient descent. 
% As optimization is not in the scope of this 
% paper, 
We refer the readers to Bengio~\cite{Bengio2012}
and references therein for details on optimization.

In essence, we can view the AE as a deterministic model 
that maps a set of inputs (features in our case) into 
their reconstruction. This is in contrast to the generative
model variants of deep neural network, such as the 
VAE and generative adversarial network (GAN), which is capable of generating new data based on the distribution of the training data.

\subsection{Variational Autoencoder (VAE)}

Unlike AE that deterministically encodes the  
inputs into their latent representation and subsequently
produce a reconstruction, the 
VAE \cite{KingmaWelling2013} 
is a generative model that treats the latent 
representation layer as random variables conditional
on the inputs. Although the encoder and decoder 
in the VAE follows the same computation model as 
the AE as in Equation~\eqref{eq:AE_computation},
the encoding process is instead used to compute the 
parameters for the \textit{conditional distributions} 
of the latent representation. 
The parameters can then be used to generate or 
sample the latent representation for decoding.
% % KarWai: I think a bit too technical, so leave for future
% \todo{[Any intuition why this is useful? I think VAE is inspired from the variational inference, which is to approximate the posterior distribution $q(Z)$ (here $q(Z|X)$) with some parameterized distribution (here is encoder net to mean,variance of gaussians). The intuition behind the latent $Z$: for example, you want to generate handwritten digit, first you think of some latent variables: which digit it is, the style I want to write (italic, bold, etc) (sampling in latent space), then you generate the digits (decoding from latent space).]}
The conditional distributions
are generally assumed to be Gaussian for real-valued nodes. 
% or Bernoulli for binary nodes. 
For example, denote
$\vec{z}_{i}$ as the value of the latent representation layer, then it can be
written as 
\begin{align}
    \vec{z}_{i} \sim \mathcal{N}(\vec{\mu}_{i}\,, \mathrm{diag}(\vec{\sigma}_{i}^2))
\label{eq:conditional_distribution_z}
\end{align}
where $\mathrm{diag}$ denotes a function that transforms
a vector into a diagonal matrix; and
$\vec{\mu}_{i}$ and $\vec{\sigma}^2_{i}$ are the mean
and variance for the conditional Gaussian distribution
obtained from the output of the encoder:
\begin{align}
    \vec{\mu}_{i} & = \mathbf{W}_{i}^\mu \, \vec{h}_{i-1} + \vec{b}_{i}^\mu \\
    \log \vec{\sigma}_{i}^2 & = \mathbf{W}_{i}^\sigma \, \vec{h}_{i-1} + \vec{b}_{i}^\sigma
\,.
\end{align}
The parameters $\mathbf{W}_{i}^\mu$, $\mathbf{W}_{i}^\sigma$,
$\vec{b}_{i}^\mu$, and $\vec{b}_{i}^\sigma$ are interpreted
in the same way as in the autoencoder above. Treatment on
the hidden layers is identical to that of autoencoder.

The probabilistic nature of the VAE also means that we 
cannot simply employ the usual learning algorithm 
on standard objective function (e.g.\ mean square error) 
to train the model.
Instead, a class of approximate statistical inference method 
are used, which is called the Variational Bayes
(thus gives rise to the name VAE).
As the discussion on the inference methods are rather 
technically involved, we refer the interested readers to 
Kingma and Welling \cite{KingmaWelling2013} for
details. Put simply, an alternative objective function
known as the \emph{variational lower bound} is optimized, and
stochastic sampling is used for approximation. 
% Note that an improvement to this method, which uses a better sampling
% scheme, has been proposed by Burda et al.~\cite{BurdaGrosseSalakhutdinov2015}. Their
% approach will be incorporated in our training algorithm.

In terms of architecture, VAE is similar to AE and is illustrated in Fig.~\ref{fig:autoencoder}.
% that is, the latent representation layer is of size 100 accompanied 
% by three hidden layers each in the encoder and 
% the decoder. Note that although the originally proposed VAE
% \cite{KingmaWelling2013} has simpler architecture, it is not
% uncommon to have more complex ones in practice, such as those
% in Rezende et al.~\cite{RezendeMohamedWierstra2014}.
% We let the latent representation to be real-valued and 
% thus assume a Gaussian conditional distribution. 
The 
ReLU activation function is used by the encoder
and the decoder in all of the intermediate layers, and
the linear activation $g(x) = x$ will be used 
for the output.

% Not needed for this paper?
% \todo{To include Conditional VAE? \cite{SohnLeeYan2015}}

\section{\GEE: Anomaly Detection Framework}
\label{sec:method}

% \begin{figure}[ht]
%     \begin{center}
%     % \vskip 0.02in
%         \includegraphics[width=0.95\columnwidth]{figs/workflow.pdf}
%         \caption{
%             Tentative Flow Chart
%         }
%         \label{fig:workflow}
%     \end{center}
%     % \vskip -0.10in
% \end{figure}

% \hlt{
Our anomaly detection framework \GEE consists of the following main steps.
First, important information from the NetFlow data, such
as the average packet size, entropy of destination ports, etc., are extracted 
to obtain a set of features. Then, the features are fed into the
VAE for learning the normal behaviour of the network. Subsequently,
anomalies are identified based on their reconstruction errors. 
% We outline each process in details in the following.
Finally, we describe how we can employ the gradient 
information available from the VAE for explanation and for
%feature fingerprinting.
fingerprinting attacks. 
% }

\subsection{Feature Extraction}
\label{subsec:feature_extraction}

A NetFlow record is a set of packets that has the same
five-tuple of source and destination IP addresses, source and destination ports, and protocol. In addition to the above, 
some of the important fields that are commonly available
in NetFlow records are start time of the 
flow (based on the first sampled packet), duration, number of packets, number of bytes, and TCP flag. 
% The NetFlow records are often collected in the time scale of seconds and is too fine-grained \todo{remove this sentence?}.
We group the NetFlow records into 3-minute sliding windows 
based on the source IP addresses to form aggregated features.
This means that each \emph{data point} in this paper 
corresponds to the network statistics 
of a single source IP address within a 3-minute period. 
Note that such extraction allows us to identify the offending IP
address and also the time window an anomaly belongs to,
which are important for further analysis and decision making.
The period of 3 minutes is chosen to balance between the
practicality and quality of the aggregated statistics, 
where the statistics will be insignificant if the period
is too short; while using a long time window means we cannot
perform real time analysis.

Overall, we extract 53 aggregated features, which include
% as inputs
% to the VAE. Extracted features include the following:
\begin{itemize}
    \item 
        mean and standard deviation of flow durations, 
        number of packets, number of bytes, packet rate; and
        byte rate;
        % (over the 3-minute duration);
    \item
        entropy of protocol type, destination IP addresses, 
        source ports, destination ports, and TCP flags; and
    \item
        proportion of ports used for common applications 
        (e.g.\ WinRPC, Telnet, DNS, SSH, HTTP, FTP, and POP3).
\end{itemize}
To ensure that meaningful statistics are captured, data point 
that contains too few flows (less than 10 in this case) 
are removed from the dataset. 
This reduces noise in the training data.
% and thus enabling the VAE model to
% learn the standard behaviour of the network without being
% corrupted by random spurious effect. 
Finally, the statistics
are either scaled to between 0 and 1 
or normalized into Z-score~\cite{JainNandakumarRoss2005} 
as input features for the VAE.

\subsection{Unsupervised Learning of VAE}

Training algorithm of the VAE is implemented using
TensorFlow
\cite{AbadiBarhamChen2016}, 
which provides
powerful tools of automatic differentiation and
comes with built-in optimization routines.
As pointed out in Section~\ref{sec:model}, the 
training of the VAE is fairly complex and beyond the scope of this paper,
so we provide only a brief outline here. Before
starting the training procedure, the parameters 
in the VAE are randomly initialized. This subsequently
allows us to perform a \emph{forward pass} on
the encoder by computing the distribution of the 
latent representation layer via 
Equation~\eqref{eq:conditional_distribution_z}. 
With this, several samples can be generated
from the Gaussian distribution which are 
used to compute the variational
lower bound, which consists of a KL divergence
term and an expectation term:
\begin{align}
    \mathcal{L} =
        - D_{KL}[q(\vec{z}\,|\,\vec{x})\,||\,p(\vec{z})] 
        + \mathbb{E}_q[\log p(\vec{x}\,|\,\vec{z})]
\end{align}
where $\vec{z}$ is the latent representation of
the input features $\vec{x}$. Here, the distribution 
$p(\cdot)$ corresponds to the Gaussian prior and conditional
distribution of the VAE model; while $q(\cdot)$ is a
variational approximation~\cite{BleiKucukelbirMcAuliffe2017}
of $p(\cdot)$, generally chosen to be Gaussian as well.
Refer to Kingma
and Welling~\cite{KingmaWelling2013} for details.
Fortunately, this objective 
function can be maximized with stochastic 
optimization techniques since the gradients
are readily available via automatic 
differentiation~\cite{BaydinPearlmutterRadulSiskind2017}.

Here, we employ Adam~\cite{KingmaBa2015}
as the optimization algorithm, which enables 
training in minibatches. Generally, real-time 
training can be achieved by choosing a small 
minibatch size and discarding the data after 
one epoch. 
We like to highlight that \emph{label information is not used at 
all during training}.

\subsection{Anomaly Detection}

Once the parameters are optimized after training,
the VAE model is used for anomaly detection,
where an IP address and its time window is
recognized as abnormal when the reconstruction 
error of its input features is high.
Here, the reconstruction error is the mean 
square difference between the observed 
features and the expectation of their reconstruction
as given by the VAE. 
A high reconstruction error is generally 
observed when the network behavior differs 
greatly from the normal behavior that was learned 
by the VAE. 
The threshold is usually selected such that 
we treat a small percentage (say 5\%) of the data
as anomalies. Otherwise, we can make use of the
labeled information to select a threshold to 
maximize the detection rate while having small
false positive rate.
An illustration of how this can be done is 
suggested in the next section, with the aid of
Fig.~\ref{fig:reconstr_error}.
Note the anomalies are associated with unusual
network behaviours from a particular source
IP address, and may not necessarily 
be malicious.

\subsection{Gradient-based Explanation for Anomalies}

While most existing anomaly detection works in the literature consider only the evaluation on detection accuracy, we go beyond and provide an explanation
on why a data point is flagged as abnormal. 
This is significant since it challenges 
the popular belief that a deep learning model 
functions as a black box 
that cannot be interpreted; a VAE model can in fact
be used to explain why an IP address is treated
as an anomaly. This is done by analyzing the 
gradients `contributed' by each feature 
of the data point, which is obtainable from the VAE through
automatic differentiation in TensorFlow.

% in practice 
% since even as small as a 5\% false positive rate results in large amount of flows being tagged as anomalous and thus overwhelming the \todo{analysts/users} with too much information.

The key question to ask that leads to our approach is:
\emph{How does the VAE's objective function vary if a feature in the anomaly data point increases or decreases by a small amount?}
Intuitively, given the trained VAE and an anomalous data point, if the function (reconstruction error) changes quite a lot when a particular feature of the anomalous data point is varied by a small amount, then this feature at its current value is significantly abnormal, since it would like to perturb the VAE model (through optimization) to fit itself better.

Gradients, or more technically the derivative of the variational lower bound, ${\partial \mathcal{L}}/{\partial f_{ij}}$\,,
% \begin{align}
%     \mathrm{Gradient} =
%     \frac{\partial \mathcal{L}}{\partial f_{ij}} \,,
% \end{align}
are computed for each feature $f_{ij}$ from each data point $i$.
Two applications of the gradient can be immediately derived.
Firstly, even without having the ground truth labels, the flagged
anomalies can be clustered based on their gradients %$\vec{f}_i$ 
into 
groups that share similar behavior, making it easier for analysts to investigate.
%For example, port scanning
%attacks may exhibit behaviors that are similar to DoS 
%(see Fig.~\ref{fig:clusterdist} in Section~\ref{sec:eval}
%for details).
Secondly, if we have the labeled information on certain types 
of attacks, then we can derive gradient-based fingerprints that
associate with the attacks. These fingerprints can be used to 
identify specific attacks from another day. Of course, the anomalies that are identified through the fingerprints are more accurate  
since labeled information was indirectly used in a way similar
to semi-supervised learning.
The anomalies are detected through the 
L2 distance computed from the normalized gradient vectors.
The rationale of using such formulae is presented 
next.

% in the 
% next section.

\section{Dataset and Evaluation}
\label{sec:eval}

\begin{table}[tb]
    \vspace{0.1in}
    \caption{
        Volume of the NetFlow records (in thousands).
    }%
    \label{tbl:data} 
    \vspace{-0.1in}
    \begin{center}
    \begin{tabular}{r
    S[table-format=3.0]
    S[table-format=3.0]
    S[table-format=3.0]
    S[table-format=3.0]
    S[table-format=3.0]
    S[table-format=3.0]
    S[table-format=3.0]
    }
        \toprule
        {Date (2016)}
        & {Total}
        & {\!DoS\!}
        & {\!B.net\!}
        & {\!Sc11\!}
        & {\!Sc44\!}
        & {\!Spam\!}
        & {\!B.list\!}
        \\
        \midrule
        \underline{Training Set} & & & & \\[0.5mm]
        Mar 19 (Sat)
        &  {110M} & {-} & {-} & {-} & {-} & 795 & 352 \\
        Jul 30 (Sat)  
        &  {110M} & 779 & 152 & 98 & 373 & {-} & 293 \\[0.1mm]
        \hdashline
        \underline{Test Set} & \rule{0pt}{1.02\normalbaselineskip} & & & \\[0.5mm]
        Mar 18 (Fri)  
        &  {40M} & {-} & {-} & {-} & {-} & 13 & 194 \\
        Mar 20 (Sun)
        &  {110M} & {-} & {-} & {-} & {-} & 795 & 352 \\
        July 31 (Sun)
        &  {105M} & 784 & 152 & 72 & 369 & {-} & 225 \\
        \bottomrule
    \end{tabular}
    \end{center}
    \vskip -0.2in
\end{table}

% \todo{Present the dataset used, data size, identify the anomalies present.}

% \todo{if we only use the UGR dataset, this paragraph can be removed... 
% We consider two datasets to assess our anomaly detection
% framework: the first one corresponds to network data captured
% in a university setting with simulated botnet attacks,
% while the second dataset is real traffic captured by an ISP.
% The first dataset is known as the CTU-13 dataset 
% \cite{CTU13} which is captured by
% the Czech Technical University in Prague, Czech Republic.
% Here, thirteen kind of real botnet traffics
% (e.g.\ DDoS, port scan, spam) are generated 
% through specific malwares and are captured with normal and 
% background traffic. These traffic are processed into NetFlow
% data with about 20M total flows, the flows attributed to
% botnets range from 0.03\% to 8.11\% within each botnet
% scenario. \todo{Which scenario did we use?}
% Anomalies are considered as traffic that comes 
% from the botnets, which are labeled as 'from-botnet' in the 
% dataset. In our evaluation, we ignore other potential anomalies 
% that are present in the data due to the lack of ground truth 
% labels.

% \todo{UGR} \cite{UGR16}
For evaluation, we use the recently published UGR16 dataset~\cite{UGR16}, which contains anonymized NetFlow traces captured from a real network of a Tier 3 ISP. The ISP provides cloud services and is used by many client companies of different sizes and markets. The UGR trace is a fairly recent and large-scale data trace that contains real background traffic from a wide range of Internet users, rather than specific traffic patterns from synthetically generated data (e.g., DARPA'98 and DARPA'99~\cite{DARPA}, UNB ISCX 2012~\cite{UNB-ISCX-2012}, UNSW-NB15~\cite{UNSW-NB15}, CTU13~\cite{CTU13}). Another publicly available Internet traffic data is from the MAWI Working Group~\cite{mawi}, but the labeled data consists of only 15-minute of traffic per day. On the other hand, UGR contains traffic for the whole day over a 4-month period. Furthermore, UGR attack traffic data is a mixture of generated attacks, labeled real attacks, and botnet attacks from controlled environment. Specifically, the labeled attacks consist of:
\begin{itemize}
\item Low-rate DoS: TCP SYN packets are sent to victims with packet of size 1280 bits and of rate 100 packets/s to port 80. The rate of the attack is sufficiently low such that the normal operation of the network is not affected.
\item Port scanning: a continuous SYN scanning to common ports of victims. There are two kinds of scanning, one-to-one scan attack (Scan11) and four-to-four (Scan44). 
\item Botnet: a simulated botnet traffic obtained from the execution of the Neris malware. This data comes from the CTU13 trace~\cite{CTU13}.
\item Spam: peaks of SMTP traffic forming a spam campaign. 
\item Blacklist: flows with IP addresses published in the public blacklists. As emphasized in UGR~\cite{UGR16}, not all traffic flows involving blacklisted IP addresses are related to attacks. 
%Besides, since blacklist may contain obsolete information, it is not a reliable information for anomalies. 
However, we include it for completeness.
\end{itemize}
Other attack labels that are available in the dataset
are ignored due to their low number of occurrence, as they appear in 
less than 10 flows in total.
Also, we like to caution that the background traffic should not be
treated as fully free of attacks, since it is likely that
some attacks have avoided detection.

We select a total of five days of UGR data for our experiments. 
Two Saturdays are used as training data while 
three other days on Friday and Sundays are chosen for testing.
The statistics for the data are presented in Table~\ref{tbl:data};
NetFlow records without any labels are the background data.
Note that the data on March 18 is collected from around 10am,
thus smaller.

\begin{figure}[t]
    \centering
    \vskip -0.05in
    \begin{subfigure}{0.41\columnwidth}
        \centering
        \includegraphics[width=0.95\textwidth]{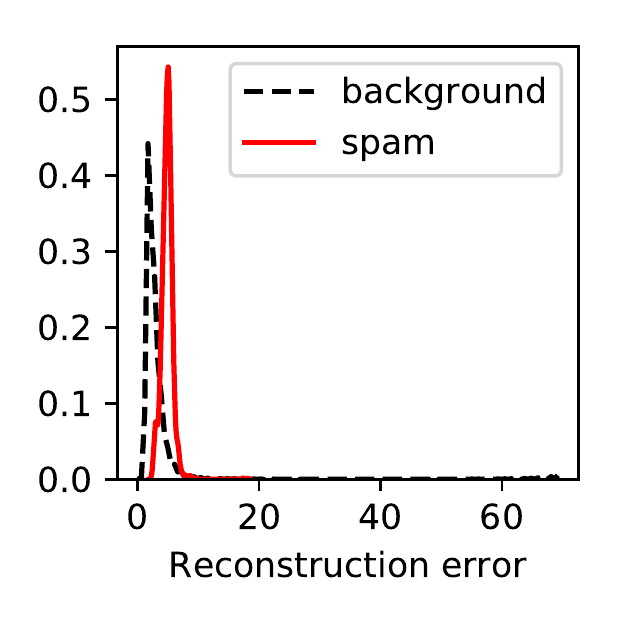}
        \caption{Spam.}
    \end{subfigure}
    \begin{subfigure}{0.41\columnwidth}
        \centering
        \includegraphics[width=0.95\textwidth]{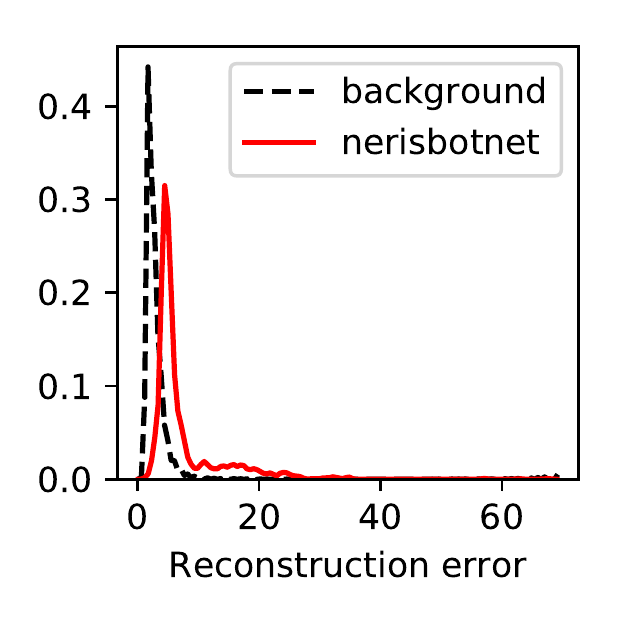}
        \caption{Botnet.}
    \end{subfigure}
    \begin{subfigure}{0.41\columnwidth}
        \centering
        \includegraphics[width=0.95\textwidth]{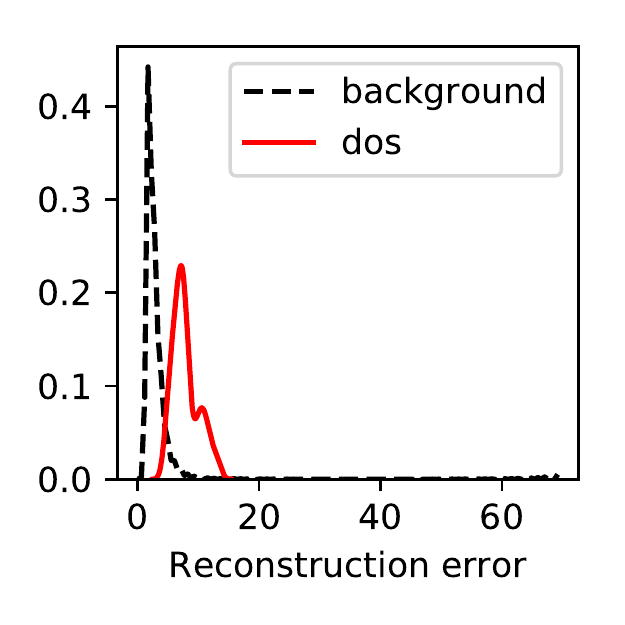}
        \caption{DoS.}
    \end{subfigure}
    \begin{subfigure}{0.41\columnwidth}
        \centering
        \includegraphics[width=0.95\textwidth]{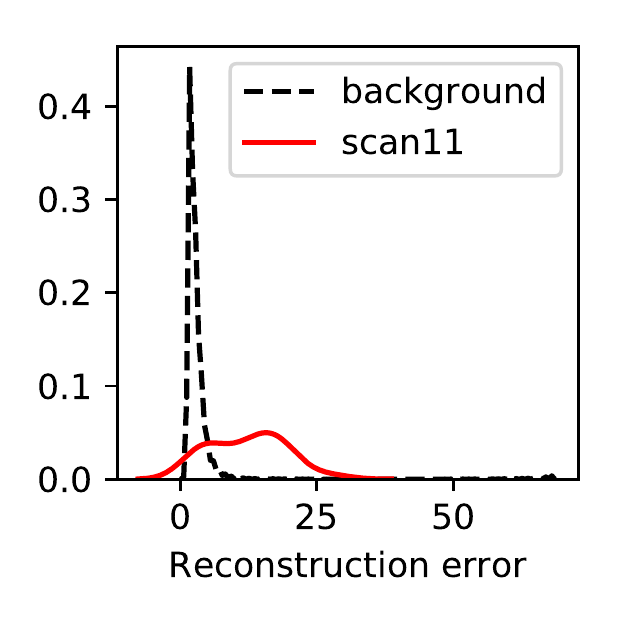}
        \caption{Scan11.}
    \end{subfigure}
    \begin{subfigure}{0.41\columnwidth}
        \centering
        \includegraphics[width=0.95\textwidth]{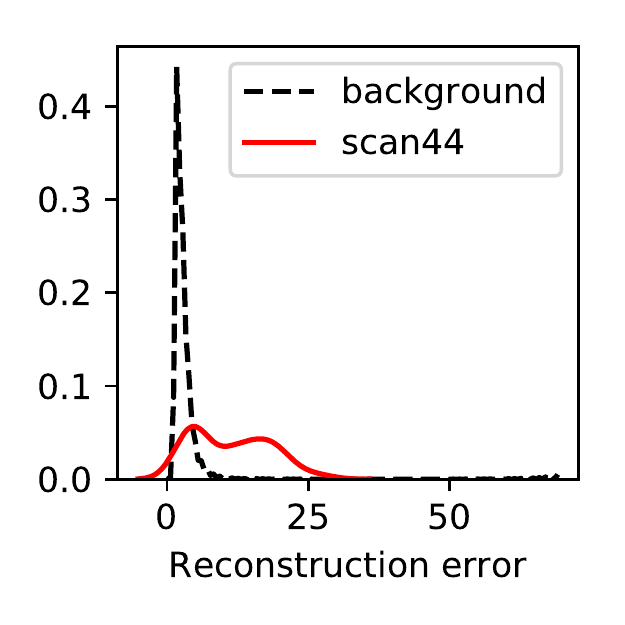}
        \caption{Scan44.}
    \end{subfigure}
    \begin{subfigure}{0.41\columnwidth}
        \centering
        \includegraphics[width=0.95\textwidth]{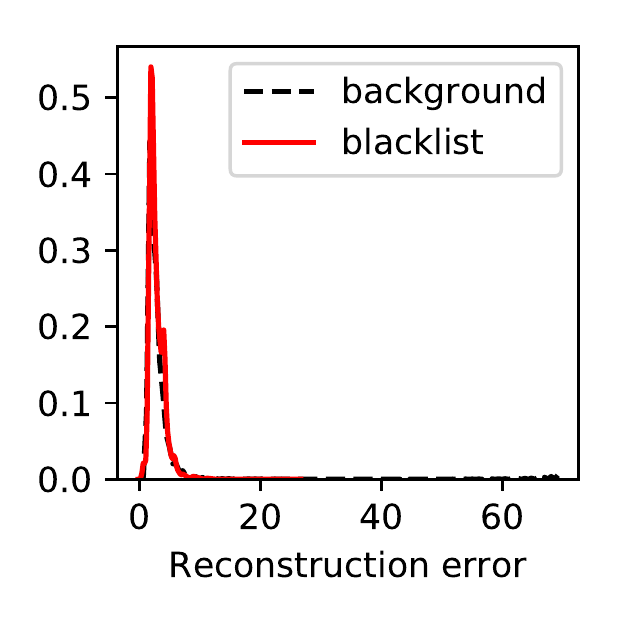}
        \caption{Blacklist.}
    \end{subfigure}
    \caption{Distribution for reconstruction error on training data.}
    \label{fig:reconstr_error}
    \vskip -0.12in
\end{figure}

After applying feature extraction as discussed in 
% Sec~\ref{subsec:feature_extraction},
Section~\ref{sec:method},
we obtain a training dataset of 5,990,295 data points.
The data are trained via stochastic optimization with 50 epochs and minibatches of size 300. 
The weight decay is set to 0.01. A brief outline of the
training procedure was given in Section~\ref{sec:method}.
For the test set, we processed a total of 
1,957,711 data points on March 18,
2,954,983 data points on March 20, and
2,878,422 data points on July 31.
For the purpose of evaluation, we say that a data point belongs
to an attack type if more than half of the flows are labeled
with such attack within the 3-minute aggregation,
otherwise it is treated as benign data.
% is greater than 50\%.

We present the distribution of the reconstruction errors
for the training data on Fig.~\ref{fig:reconstr_error}. The ground truth
labels are used to separate the anomalies from the background flows,
which let us examine whether the anomalies behave differently 
from the normal behaviors. Overall, there is some overlap with the
background for spam, botnet, DoS, and scanning activities, but we 
can find a cut off point to roughly separate them.
For blacklists, however, their behaviors are indistinguishable from
the background traffic. 
% \todo{What is the explanation here?}

\begin{figure}[t]
    \centering
    \vskip -0.05in
    \begin{subfigure}{0.42\columnwidth}
        \centering
        \includegraphics[width=0.95\textwidth]{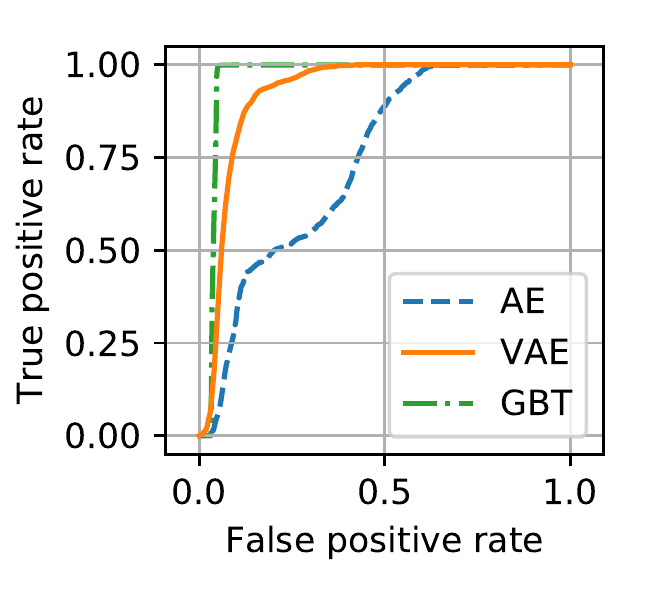}
        \caption{Spam.}
    \end{subfigure}
    \begin{subfigure}{0.42\columnwidth}
        \centering
        \includegraphics[width=0.95\textwidth]{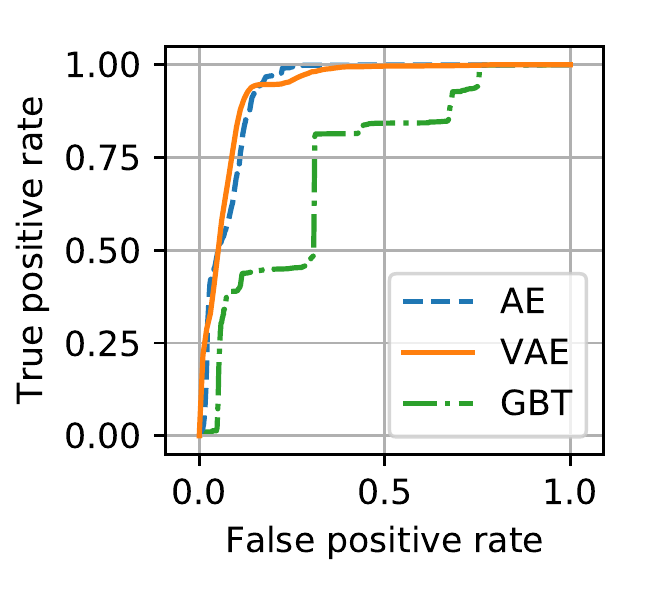}
        \caption{Botnet.}
    \end{subfigure}
    \begin{subfigure}{0.42\columnwidth}
        \centering
        \includegraphics[width=0.95\textwidth]{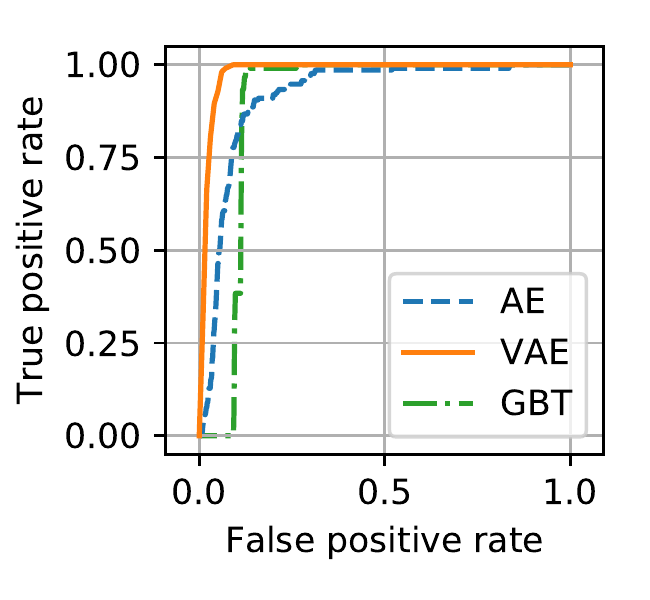}
        \caption{DoS.}
    \end{subfigure}
    \begin{subfigure}{0.42\columnwidth}
        \centering
        \includegraphics[width=0.95\textwidth]{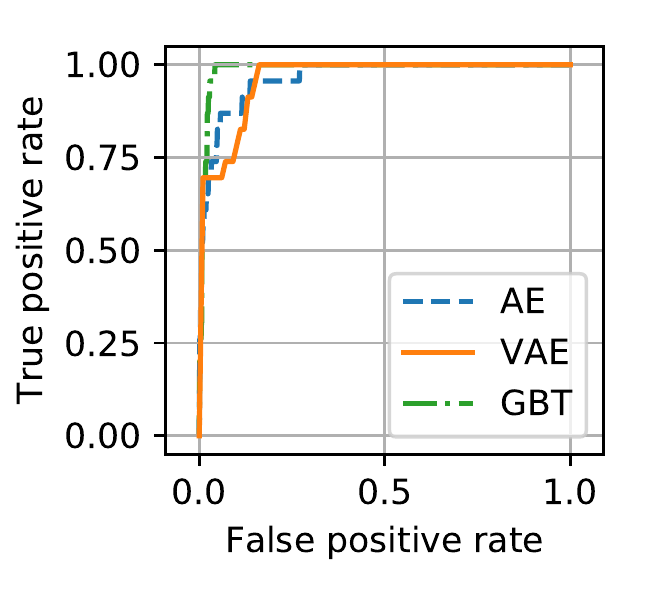}
        \caption{Scan11.}
    \end{subfigure}
    \begin{subfigure}{0.42\columnwidth}
        \centering
        \includegraphics[width=0.95\textwidth]{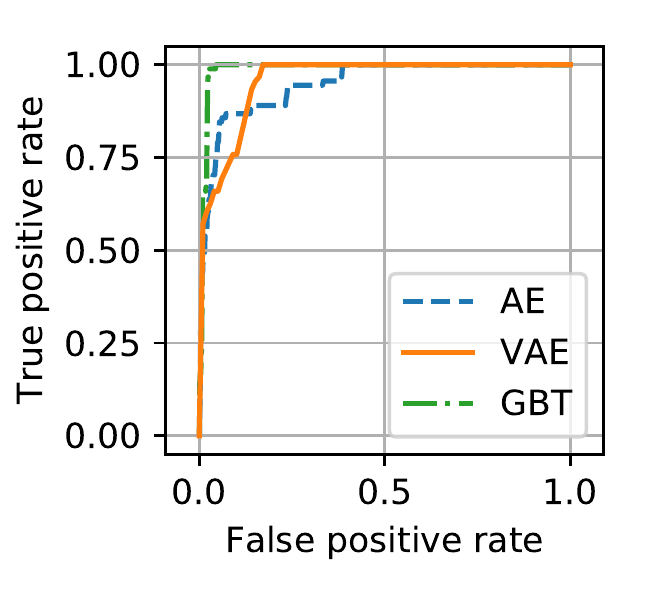}
        \caption{Scan44.}
    \end{subfigure}
    \begin{subfigure}{0.42\columnwidth}
        \centering
        \includegraphics[width=0.95\textwidth]{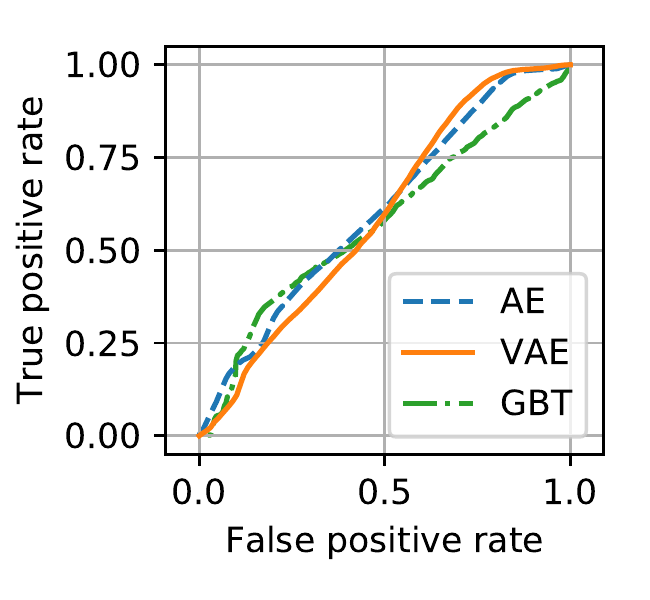}
        \caption{Blacklist.}
        \label{subfig:ROC_blacklist}
    \end{subfigure}
    \caption{ROC for VAE, AE, and GBT on training data..}
    \label{fig:roc1train}
    \vskip -0.10in
\end{figure}

\subsection{Baseline}

We compare our proposed anomaly detection framework \GEE 
that uses VAE against that of AE and also a 
\emph{Gaussian Based Thresholding} (GBT) approach. 
% [Future work] Since the AE might overfit to the training data and 
% subsequently also learn to reconstruct the anomalies 
% in the data, we consider an additional evaluation in which
% the AE `cheats' by removing the anomalies (by looking at the 
% ground truth labels) from the training data. 
For a fair comparison, the baseline AE shares the same 
architecture as the VAE, and as illustrated in Fig.~\ref{fig:autoencoder}.
The AE is implemented using Keras,
% \footnote{\url{https://keras.io/}} 
a high level open source neural network library. We use the
same TensorFlow backend for training the AE. This is by
minimizing the reconstruction error (mean square error) using the stochastic 
optimizer Adam~\cite{KingmaBa2015}, with minibatch size
chosen to be 256. Similar to the VAE, data points that 
have large reconstruction error are flagged as anomalies.

For the GBT, we fit independent but
non-identical Gaussian distribution models to the features
to learn the standard behaviors of the data. Then, we 
compute the Z-score for all features in the testing 
dataset and use the product of the average, standard deviation,
and the maximum of the Z-scores as final score for
anomaly detection. The data point with score that exceeds a certain
threshold is considered as anomaly. 
We emphasize that both the AE and GBT are trained on the same
extracted features as in VAE, this is to ensure the comparison
is performed over the models rather than feature engineering.
% Similar to AE, we consider
% two versions of this method: 1) one where the training data 
% is the full traffic, and 2) one with labeled anomalies 
% removed.

\subsection{ROC Performance on Anomaly Detection}

We evaluate the performance of these three methods (VAE, AE and GBT) using 
Receiver Operating Characteristic (ROC) curves, where
the true positive rates are plotted against the 
false positive rates by varying the sensitivity or threshold
for anomaly detection. A good performance is where 
the true positive is high while the false positive is low.

\begin{figure*}[t!]
    \vspace{-2mm}
    \centering
    \includegraphics[width=0.7\textwidth]{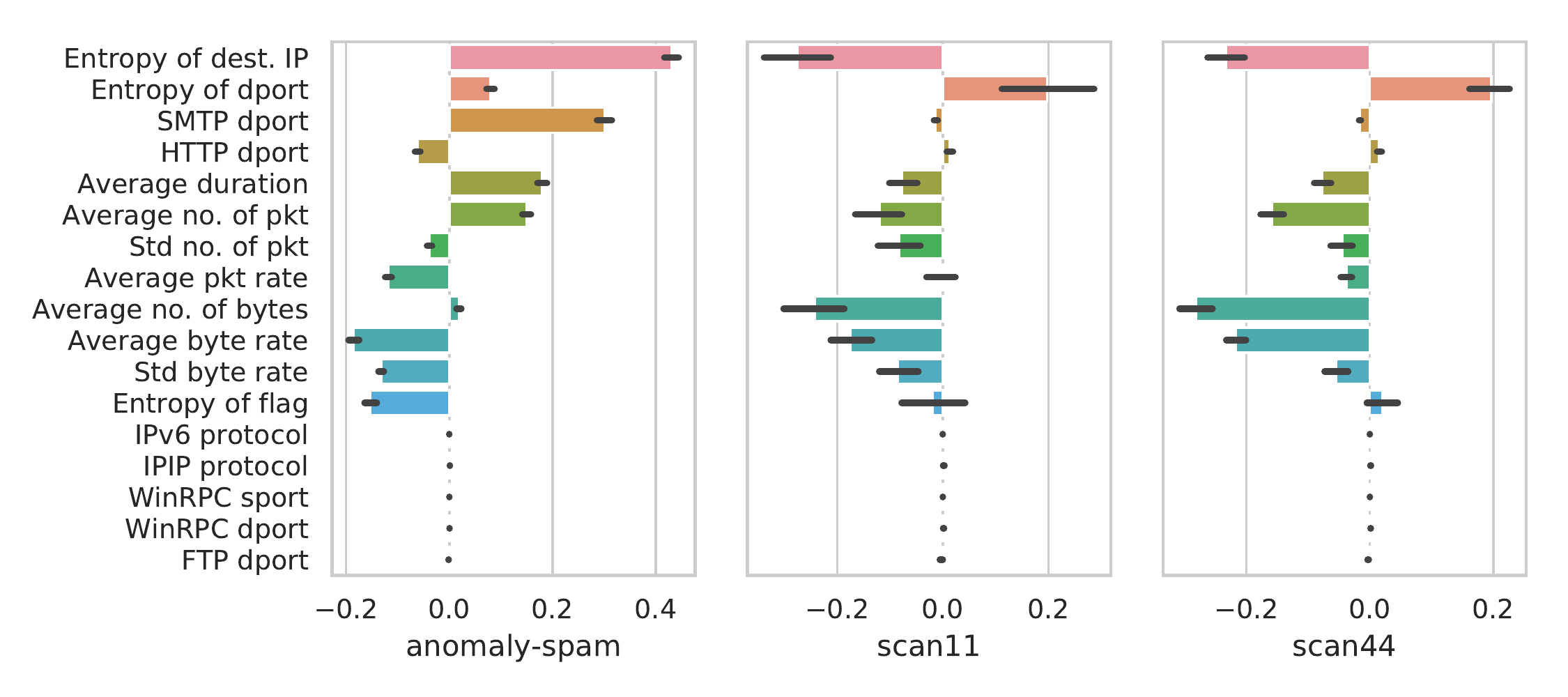}
    \caption{Normalized gradients of spam, scan11, and scan44 on selected features.}
    \label{fig:traingrad}
    \vskip -0.10in
\end{figure*}

\begin{figure}[t]
    \centering
    \vskip -0.05in
    \begin{subfigure}{0.42\columnwidth}
        \centering
        \includegraphics[width=0.95\textwidth]{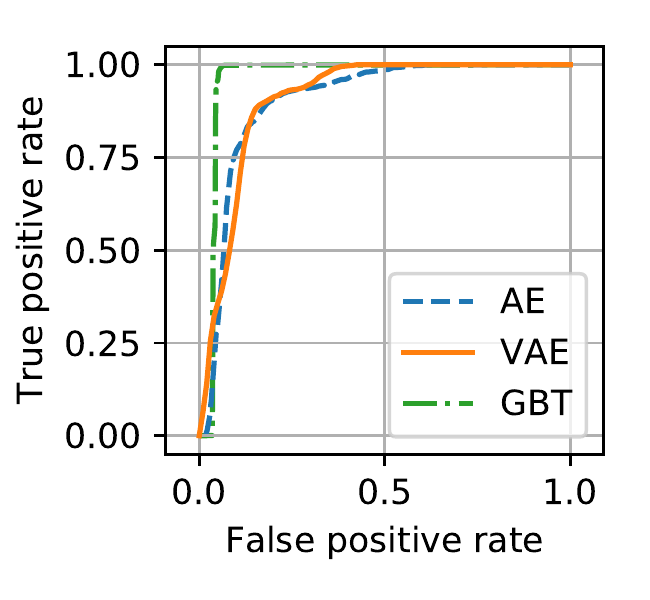}
        \caption{Spam on March 20.}
    \end{subfigure}
    % \begin{subfigure}{0.42\columnwidth}
    %     \centering
    %     \includegraphics[width=0.95\textwidth]{figs/test/mar18/ROC/ROC_spam_vae_ae_thres.pdf}
    %     \caption{Spam on March 18.}
    % \end{subfigure}
    \begin{subfigure}{0.42\columnwidth}
        \centering
        \includegraphics[width=0.95\textwidth]{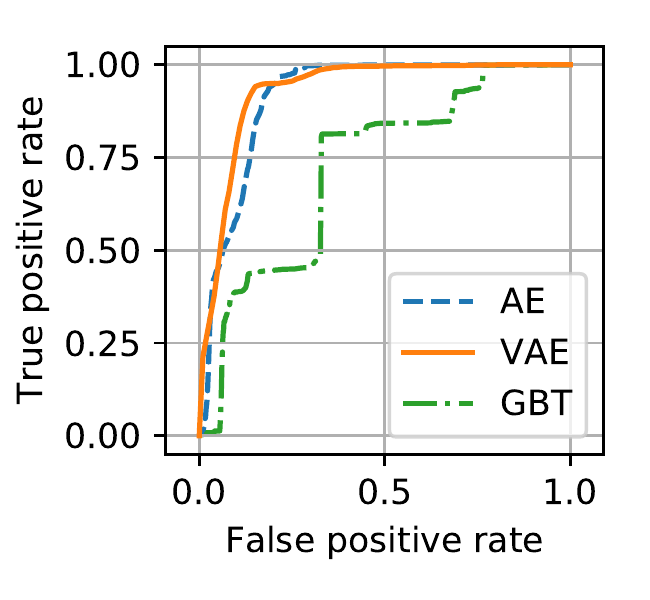}
        \caption{Botnet on July 31.}
    \end{subfigure}
    \begin{subfigure}{0.42\columnwidth}
        \centering
        \includegraphics[width=0.95\textwidth]{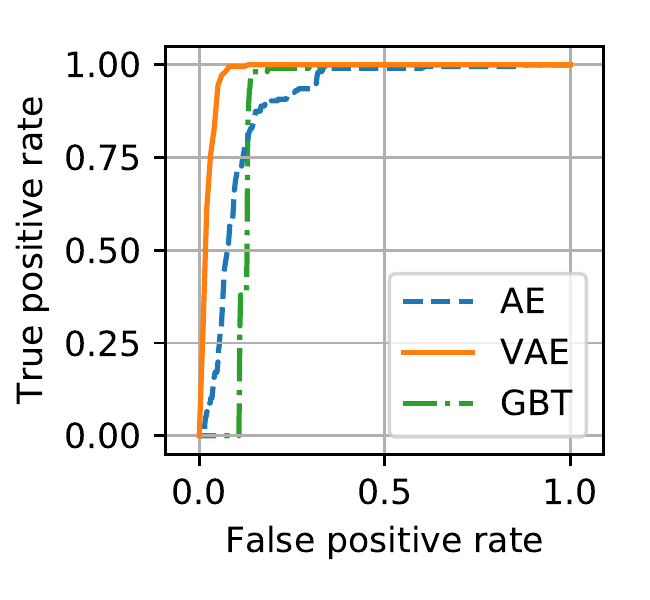}
        \caption{DoS on July 31.}
    \end{subfigure}
    % \begin{subfigure}{0.42\columnwidth}
    %     \centering
    %     \includegraphics[width=0.95\textwidth]{figs/test/jul31/ROC/ROC_scan11_vae_ae_thres.pdf}
    %     \caption{Scan11 on July 31.}
    % \end{subfigure}
    \begin{subfigure}{0.42\columnwidth}
        \centering
        \includegraphics[width=0.95\textwidth]{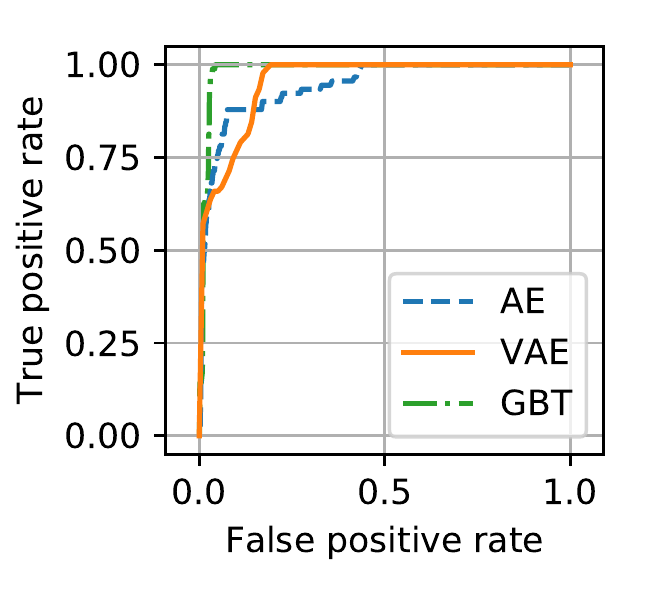}
        \caption{Scan44 on July 31.}
    \end{subfigure}
    \caption{ROC for VAE, AE, and GBT on test dataset.}
    \label{fig:roc1test}
    \vskip -0.13in
\end{figure}

We plot the ROC results on the training dataset in Fig.~\ref{fig:roc1train}
and on the testing dataset in Fig.~\ref{fig:roc1test}. The results on the training data are presented because unlike supervised learning, the labeled data is not used for training, hence the results indicate how good the algorithms are in finding anomalies from statistics of the same day. On the other hand, results for the testing data evaluate on how the anomaly detection generalizes to a new day 
based on statistics from another day in the training data. For reference, we include the results for blacklisted IP addresses which clearly cannot be identified through the statistics from the NetFlow records, see Fig.~\ref{subfig:ROC_blacklist}.

From the results, we make the following observations. 
First, threshold based approaches such as GBT work well for attacks that increase the volume of certain categories of traffic significantly, such as spam and port scanning. However, such approaches do not work as well for botnet and low rate DoS. On the other hand, AE does not work well for spam and low rate DoS. For these two attacks, AE may be unable to differentiate spam from regular email traffic because of the high volume, and unable to detect DoS due to the low rate. Overall, VAE is the most robust and has good performance on all attack types. Table~\ref{tbl:auc} summarizes the individual and average area under the ROC curve (AUC) for various attacks (except blacklist) and detection models. Clearly, VAE has the best overall performance in terms of AUC. 
Note that the AUC results on the test set are aggregated over the different days,
i.e., spam attacks are from March 18 and 20, while DoS, botnet, and port scanning
attacks are originated from July 31.

\begin{table}[bt]
    \vspace{0.05in}
    \caption{
        Area under the curve (AUC) of the ROC.
    }%
    \label{tbl:auc} 
    \vspace{-0.1in}
    \begin{center}
    \begin{tabular}{r
    S[table-format=1.3]
    S[table-format=1.3]
    S[table-format=1.3]
    S[table-format=1.3]
    S[table-format=1.3]
    S[table-format=1.3]
    }
        \toprule
        {Model}
        & {DoS}
        & {BotNet}
        & {Scan11}
        & {Scan44}
        & {Spam}
        & {Average}
        \\
        \midrule
        \underline{Train} & & & & \\[0.5mm]
        GBT
        &  0.892 & 0.735 & 0.989 & 0.989 & 0.961 & 0.913 \\
        AE      
        &  0.917 & 0.931 & 0.965 & 0.948 & 0.741 & 0.900 \\
        VAE    
        &  0.986 & 0.940 & 0.966 & 0.961 & 0.927 & {$\textbf{0.956}$} \\[0.1mm]
        \hdashline
        \underline{Test} & \rule{0pt}{1.04\normalbaselineskip} & & & \\[0.5mm]
        GBT  
        &  0.875 & 0.722 & 0.980 & 0.985 & 0.970 & 0.906 \\
        AE
        &  0.898 & 0.915 & 0.961 & 0.945 & 0.791 & 0.902 \\
        VAE
        &  0.983 & 0.936 & 0.949 & 0.958 & 0.908 & {$\textbf{0.947}$} \\
        \hdashline
        \rule{0pt}{1.02\normalbaselineskip}F.print
        &  0.980 & 0.914 & 0.976 & 0.977 & 0.965 & {$\textbf{0.962}$} \\
        \bottomrule
    \end{tabular}
    \end{center}
    \vspace{-0.2in}
\end{table}

\subsection{Identifying and Using Gradient Fingerprints}

In this section, we compute and identify the gradients for all the features for various attacks. Fig.~\ref{fig:traingrad} shows the gradients of the features for spam, scan11, and scan44. Recall that 53 features are used as input to VAE and significant features with sufficient variation in gradients are shown in the figure. The black bars reflect one standard error for 
the gradients, which are useful for assessing the significance
of the gradients, i.e., whether it is due to noise or not. 
We also present the features that are not significant for
contrast.
Note that similar gradient-based fingerprints are available 
for the other labeled attacks, not shown here due to space.
% Overall, we are able to identify such gradient based fingerprints for all the labeled attacks in the dataset, but due to space limitation, we are not able to show all of them. 

Based on the results, we can make the following observations. First, only a small subset of the features have large gradients. Second, these features with the greatest absolute gradients provide an explanation for why these flows of an IP are detected as anomalies. For example, in the case of spam attacks (which includes sending spam emails), five features have more positive gradient (higher than the learned normal) while four features have much negative gradient (lower than learned normal). Next, these combination of gradient and features can be used as a fingerprint to identify or cluster similar attacks. For example, it can be observed from Fig.~\ref{fig:traingrad} that Scan11 and Scan44 have similar gradient fingerprints.

\begin{figure}[t]
    \centering
    \begin{subfigure}{0.41\columnwidth}
        \centering
        \includegraphics[width=0.99\textwidth]{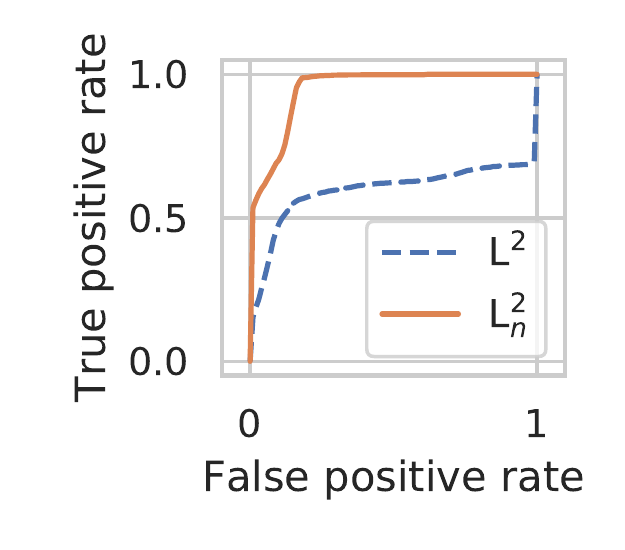}
        \caption{Spam on March 20.}
    \end{subfigure}
    % \begin{subfigure}{0.41\columnwidth}
    %     \centering
    %     \includegraphics[width=0.99\textwidth]{figs/test/mar18/gradROC/anomaly-spam_detection_05_r0_t0.pdf}
    %     \caption{Spam on March 18.}
    % \end{subfigure}
    \begin{subfigure}{0.41\columnwidth}
        \centering
        \includegraphics[width=0.99\textwidth]{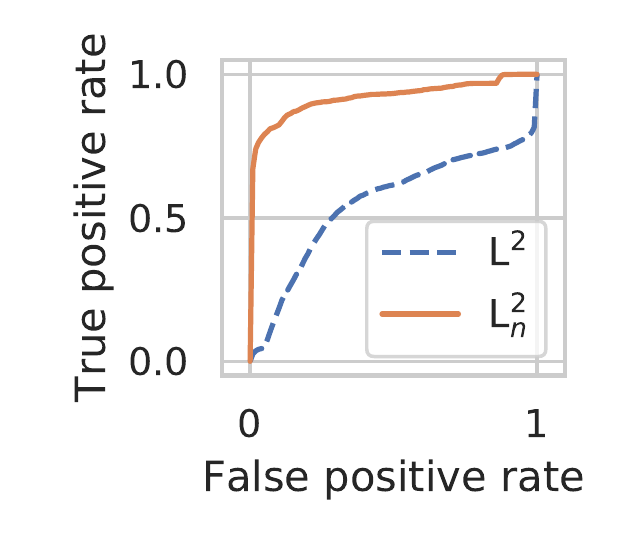}
        \caption{Botnet on July 31.}
        \label{subfig:ROC_fingerprint_botnet}
    \end{subfigure}
    \begin{subfigure}{0.41\columnwidth}
        \centering
        \includegraphics[width=0.99\textwidth]{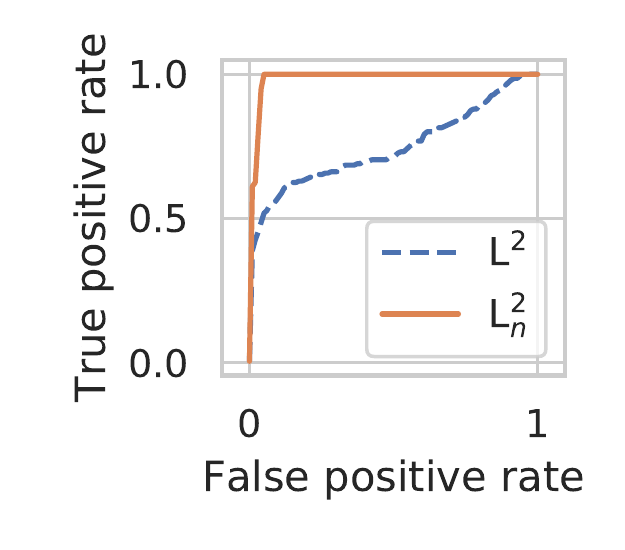}
        \caption{DoS on July 31.}
    \end{subfigure}
    % \begin{subfigure}{0.41\columnwidth}
    %     \centering
    %     \includegraphics[width=0.99\textwidth]{figs/test/jul31/gradROC/scan11_detection_05_r0_t0.pdf}
    %     \caption{Scan11 on July 31.}
    % \end{subfigure}
    \begin{subfigure}{0.41\columnwidth}
        \centering
        \includegraphics[width=0.99\textwidth]{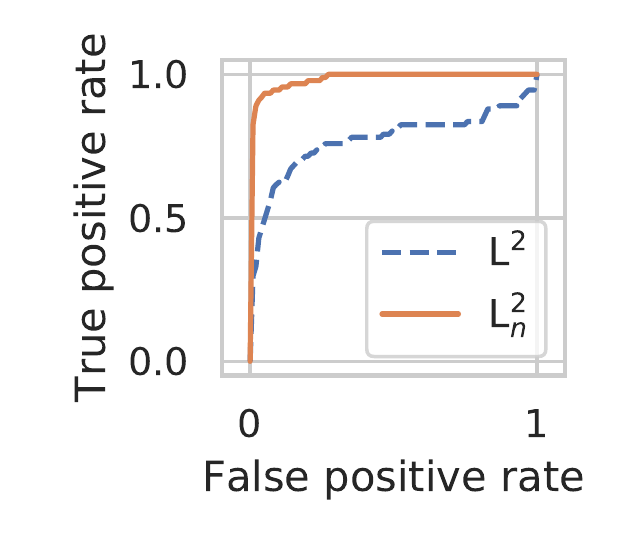}
        \caption{Scan44 on July 31.}
    \end{subfigure}
    \caption{ROC for anomaly detection using fingerprints.}
    \label{fig:roc2test}
    \vskip -0.10in
\end{figure}

To further validate our claim that these gradient fingerprints are useful in identifying similar attacks, we plot the ROC for various attacks detected in the following way. 
First, let $L^2$ denote the Euclidean distance between the average gradient fingerprint obtained from data with labeled attacks and the gradients of the VAE's objective function w.r.t.\ each test data point. Similarly, define $L^2_n$ as the same distance metric but computed on normalized gradient vectors.
%we compute the $L^2$ distance and the $L^2$ normalized gradients (denoted as $L^2_n$) between the gradient fingerprint learnt from labeled data to the gradients of the VAE loss w.r.t every point in the dataset. 
The ROC is then produced by varying the threshold on $L^2$ or $L^2_n$ distance. If the gradient fingerprint is a good measure of an attack, we expect the ROC to have high true positive and low false positive. 

The results are shown in Fig.~\ref{fig:roc2test}. It is encouraging to see that with the use of $L^2_n$, the gradient fingerprints learned are indeed good representations of these attacks. In fact, it is an improvement upon using the reconstruction error for anomaly detection, see the last row on Table~\ref{tbl:auc} for the its AUC.
We also note that the result is slightly worse for botnet, see Fig.~\ref{subfig:ROC_fingerprint_botnet}. This could be because there are many patterns behind the botnet anomalies. Hence, simply averaging all normalized gradients for botnet might not be the best approach. 
% An improvement may be to represent botnet with multiple fingerprints by clustering the botnet normalized gradients.

Finally, we may also conclude that we have discovered a new attack if an unseen gradient fingerprint is identified.

\subsection{Clustering of Anomalies}

% It is important to separate the anomalies into different
% groups based on the type of attacks, say distinguishing SSH scanning
% from DDOS attacks. 

Another possible use of the VAE generated gradients is that they can be used to cluster or group the different attacks. The idea is that if the clustering is effective, attacks should be limited to a relatively small number of clusters. 

\begin{figure}[t]
    \centering
    \begin{subfigure}{0.40\columnwidth}
        \centering
        \includegraphics[width=0.95\textwidth]{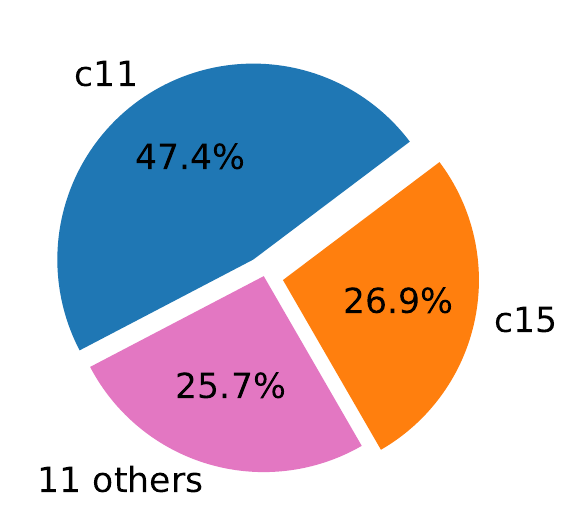}
        \caption{Spam.}
    \end{subfigure}
    \begin{subfigure}{0.40\columnwidth}
        \centering
        \includegraphics[width=0.95\textwidth]{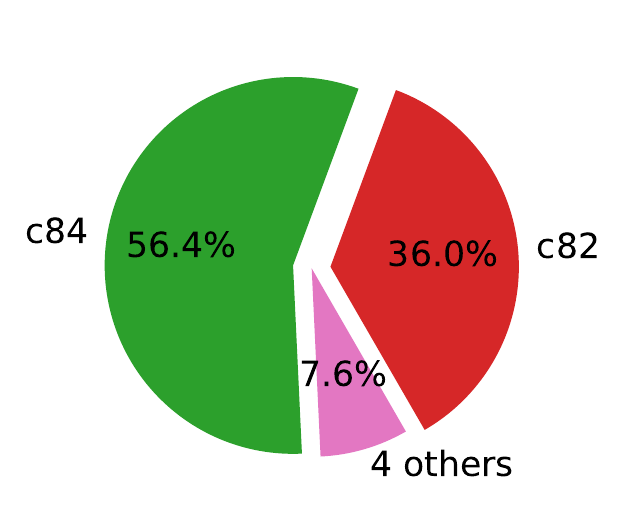}
        \caption{DoS.}
    \end{subfigure}
    % \begin{subfigure}{0.42\columnwidth}
    %     \centering
    %     \includegraphics[width=0.95\textwidth]{figs/train/clusters/attackstat/label_100_blacklist.pdf}
    %     \caption{Blacklist.}
    % \end{subfigure}
    \begin{subfigure}{0.40\columnwidth}
        \centering
        \includegraphics[width=0.95\textwidth]{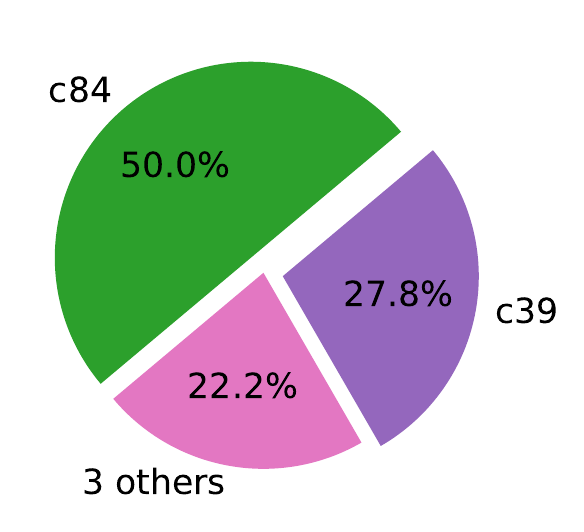}
        \caption{Scan11.}
    \end{subfigure}
    \begin{subfigure}{0.40\columnwidth}
        \centering
        \includegraphics[width=0.95\textwidth]{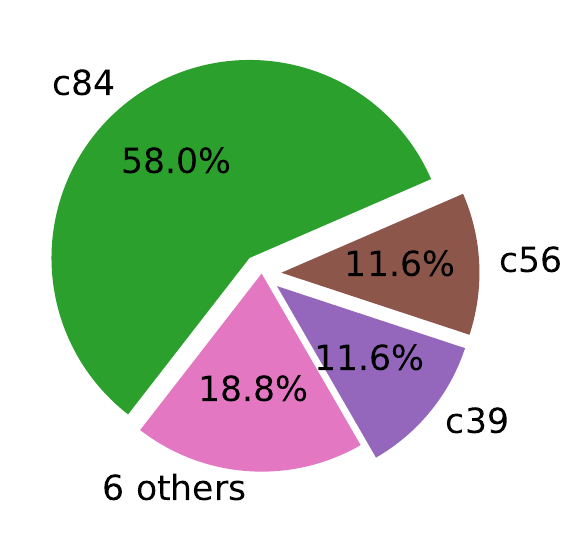}
        \caption{Scan44.}
    \end{subfigure}
    \caption{Distribution of clusters for each attack type. 
    % when clustering with k-mean of 100 clusters.
    }
    \label{fig:clusterdist}
    \vskip -0.10in
\end{figure}

We performed k-mean clustering with random initial seed on the training dataset with $k=100$. 
Fig.~\ref{fig:clusterdist} illustrates the clusters associated with the attacks discussed above. 
We find that 92.4\% of the DoS attacks appear only in two clusters (c82 and c84) with the other 7.6\% appearing in four other clusters. For spam attacks, 74.3\% of them appear in two clusters (c11 and c15), while 25.7\% appears in another 11 clusters. 
This suggests that the attacks generally exhibit small number of main behavioral patterns and thus the  
analysts can focus on a small subset of clusters to study a particular type of attacks. 
% \todo{should we mention about exploring new attacks?}

% If no ground truth label is available, or if we want to explore new type of attack in the anomalies detected from VAE, one possible direction is to do clustering on the anomaly gradient vectors (for example, we define anomaly from VAE training as data points with the top 10\% reconstruction loss). The intuition are: (1) we can interpret the anomaly based on the gradient vector as discussed in the previous section; (2) instead of looking at a large number of anomaly data points, we can look at a small number of clusters' summaries of their gradients, which says something about their behaviors. The clustering method should based on the L2 distance between normalized gradient vectors (which we illustrate in Fig.~\ref{fig:roc2test} that this is a good measure). Unfortunately, when we use simple k-mean clustering on the training data, the data points of the same attack type does not form 1 cluster. However, the good news is that majority of them belongs to only a small number of clusters, which means their behaviors are indeed clustered. For example, even we have 100 clusters, all attacks only belong to a small number of clusters (Fig.~\ref{fig:clusterdist}). However, some attacks might stay in the same cluster together. 
% Interestingly, cluster 13 shows some hints about spaming attack (Fig.~\ref{fig:spamclusterdist}).

% \begin{figure*}[ht]
%     \centering
%     \includegraphics[width=18cm]{figs/train/clusters/gradients/unit_length_axis1_ncluster100_gradient_13_barplot_cut.pdf}
%     \caption{Cluster 13 gradient vector.}
%     \label{fig:spamclusterdist}
% \end{figure*}

\subsection{Processing Overhead of VAE}

Training of the VAE is performed on a single graphical processing unit (GPU) 
{GTX 1080Ti} with Intel Xeon CPU E5-2683 (16 cores, 2.1GHz), and 256Gb of RAM, which enables
real time online training on streaming data. 
% Here, we describe the running time needed for feature extraction, minibatch training,
% and testing for anomalies.
Since the features are aggregated into a 3-minute sliding window, 
the running statistics from the NetFlow records are kept for three
minutes before discarded. On average, feature extraction took about
25 seconds for every 3-minute window of the UGR data, which corresponds
to about 200k NetFlow records. 
Note that our implementation
did not optimize for such operation, and speed improvement can easily be achieved.

Training using the TensorFlow framework is highly parallelizable and
thus runs very quickly. For every minibatch of 300 data points of aggregated
features, only 10ms is needed for training; while testing the data 
points for abnormality takes about 20ms. 
Hence, almost all the processing is due to the feature extraction. % rather than the training and testing. 
With the current setup, even without speeding up the feature extraction further, our algorithm can easily perform anomaly detection in real-time.

% It is longer for testing due to the need to compute reconstruction error and gradients. 
% The minibatch is equivalent to around 30 seconds of NetFlow records, which
% are \todo{XXX flows}.
% \todo{How many flows in 3 minutes?} 

% [KarWai: Seems too technical?]
% For train VAE well, it requires randomized minibatch to avoid the time-correlation between data points (stochastic gradient ascending assumes i.i.d. training data).

\section{Conclusion}
\label{sec:conclusion}

% \todo{Conclusion here}

We proposed \GEE, a VAE-based framework that is  
robust in detecting a variety of network based
anomalies. We observed that the VAE
performs better overall compared to the AE and
a Gaussian based thresholding (GBT) method. 
% The GBT performs best in detecting attacks that are high
% volume but suboptimal for botnet and low rate DoS.
Further, we demonstrated how to use the 
gradient information from the VAE to provide an 
explanation for the flagged 
anomalies. Also, the gradient-based
fingerprints, when used directly for anomaly detection,
was shown to achieve an overall better performance.
% In fact,
% we find that the fingerprints are more useful
% in detecting anomalies from different days,
% at least in terms of AUC of the ROC.

A potential future research
is to use the conditional VAE (CVAE)~\cite{SohnLeeYan2015} 
for anomaly detection. The CVAE was developed
by the deep learning community as an extension of the 
VAE to allow for additional auxiliary labels that are available.
With CVAE, one can consider training an anomaly detection
system from multiple data sources that have different behaviors.

% by using the source label as conditioning variable. 
% It would also be interesting to investigate if our anomaly detection 
% framework can be deployed to new unseen dataset that may have
% overlapping behaviours with the existing one.

\section*{Acknowledgment}
This research is supported by the National Research Foundation, Prime Minister’s Office, Singapore under its Corporate Laboratory@University Scheme, National University of Singapore, and Singapore Telecommunications Ltd.

%\clearpage
% \newpage
\bibliographystyle{IEEEtran}
\bibliography{bibliography}

\phantom{.}  % this is to make the last citation aligned at the end of the page

\end{document}